# Identifying the sources of ideological bias in GPT models through linguistic variation in output

Christina P. Walker[1]    Joan C. Timoneda[2]

**Abstract**

Extant work shows that generative AI models such as GPT-3.5 and 4 perpetuate social stereotypes and biases. One concerning but less explored source of bias is ideology. Do GPT models take ideological stances on politically sensitive topics? In this article, we provide an original approach to identifying ideological bias in generative models, showing that bias can stem from both the training data and the filtering algorithm. We leverage linguistic variation in countries with contrasting political attitudes to evaluate bias in average GPT responses to sensitive political topics in those languages. First, we find that GPT output is more conservative in languages that map well onto conservative societies (i.e., Polish), and more liberal in languages used uniquely in liberal societies (i.e., Swedish). This result provides strong evidence of training data bias in GPT models. Second, differences across languages observed in GPT-3.5 persist in GPT-4, even though GPT-4 is significantly more liberal due to OpenAI's filtering policy. Our main takeaway is that generative model training must focus on high-quality, curated datasets to reduce bias, even if it entails a compromise in training data size. Filtering responses *after* training only introduces new biases and does not remove the underlying training biases.

## 1   Introduction

GPT-3.5 and 4 are increasingly popular among scholars to generate data, classify text, and complement human coders. Their black-box nature, however, has raised concerns about bias in model output, which in turn has led to a burgeoning debate around the politics of artificial intelligence (AI) and how to regulate generative models. In this article, we identify ideological biases in GPT-3.5 and 4 through a novel approach that matches model output to known linguistic and issue-based differences across countries. If biases exist, GPT-3.5 and 4 will reflect the predominant political attitudes of those who produced the training text. In countries where society is more conservative

[1] PhD Candidate, Political Science, Purdue University, West Lafayette, 47906, Indiana, USA

[2] Assistant Professor, Political Science, Purdue University, West Lafayette, 47906, Indiana, USA. timoneda@purdue.edu

(liberal), GPT models will produce more conservative (liberal) output. Moreover, OpenAI, the company that developed and owns these models, heavily filters the GPT-4 API to reduce output bias, but it does not filter the GPT-3.5 complete API[1]. This gives us an opportunity to also identify bias *across* OpenAI models, and disentangle biases stemming from the training data from those that derive from the algorithm or filters.

We focus our analysis on two political issues that are linguistically and geographically constrained: abortion and Catalan independence. For abortion, we draw text data from GPT-3.5 and 4 in Swedish, Polish, and English. In Poland, society tends to be socially conservative, while Sweden is more progressive[2,3]. Because training data in these two languages comes almost exclusively from their respective countries, we expect GPT responses to reflect more conservative views of abortion in Poland and more liberal ones in Sweden. We use English output on abortion primarily to test the full extent of OpenAI's filtering efforts, which have been concentrated on English text[4,5]. For Catalan independence, we draw data in Catalan and Spanish. Because Catalan society is, on the whole, more pro-independence than Spanish society[6], we expect GPT responses in Catalan to be more positive toward independence than responses in Spanish (within the Spanish-speaking world, Catalan independence is only a politically salient and divisive issue in Spain[6]). Therefore, through these two issues, and by tapping into *languages* and *issues* that are geographically confined, we can identify whether (1) GPT output reflects ideological biases in the training data and (2) OpenAI's filtering fixes these biases or induces new ones. Using multilevel modeling to identify significant differences in outputs, we specify two distinct types of biases: *training* and *algorithmic*. We provide novel evidence on ideological biases in OpenAI's GPT-3.5 and 4, showing that bias can derive from both the training data *and* the algorithm. Our findings regarding these two sources of bias have major implications for the politics of AI, the training and regulation of generative models, and applied researchers looking to use these models in downstream analyses.

## 2 GPT Models and Ideological Bias

Testing for ideological biases in GPT-3.5 and 4 is especially relevant because a growing number of articles use these models in measurement and downstream tasks[7–14]. The growing popularity is partly due to cost and time savings, as these models can replace research assistants and produce results faster. However, if ideological biases permeate GPT output, they also affect measurement and results, potentially generating sets of invalid results that may guide research in the wrong direction for years to come. Further, understanding the underlying ideological bias in language models is important as it can influence individuals' political behavior and decision-making[15], shaping how individuals gather information and perceive political events, policies, and candidates[16].

Despite its importance, investigating bias in and across GPT models is more difficult because they are not open source, unlike other LLMs such as BERT, RoBERTa, or LLaMA[17]. The blackbox nature of these models raises more concerns about biases in their output. Multiple studies have shown GPT-3 can generate harmful outputs linked to ideas of gender, race, and ideology, perpetuating various stereotypes[18–20]. For example, LLMs are 3 to 6 times more likely to choose an occupation that stereotypically aligns with a person's gender[21] and produce more violent outputs when the prompt includes a reference to Muslims over Christians or Hindus[18]. The prevailing hypothesis to explain output bias is that GPT text is bound to reflect the social biases in the training data, which is vast, unlabelled, and drawn from all types of online sources[22]. Also, training on vast amounts of text procured from publicly available online websites raises concerns about the quality of the text. It is likely that models learn biased patterns from the data. For example, GPT3.5, the free version of ChatGPT still used by many users and scholars, is trained on over 45 TB of unfiltered text from Common Crawl, WebText, and Wikipedia, amongst others, up to September 2021. The company then filtered the data to 570 GB to train the model[23]. Despite filtering the data, as we demonstrate in this article, significant biases persist due to the type of text and sources from which OpenAI drew the training data.

OpenAI has worked to mitigate these biases in GPT-4, the more powerful, paid version of ChatGPT, which has a broader knowledge base and enhanced safety and alignment features, making it 40% more likely to produce accurate factual responses than GPT-3.5[24]. It also incorporates a new filtering policy, intimately related to the growing literature on the politics and regulation of AI[25,26], adding sophisticated filters aimed at reducing strongly worded, biased responses common in GPT-3 and 3.5[27]. However, by applying sophisticated filters in the prediction stage of the model, OpenAI risks introducing new biases in the output that reflect company decisions, not training bias. Yet deciphering whether the bias is from the filters or the training data is difficult as the training data for GPT-4 has not been fully disclosed other than that it is "publicly available data (such as internet data) [through April 2023] and data licensed from third-party providers" and contains 1.76M parameters, improving upon GPT-3.5's 175 billion[24,27,28].

Few works have developed methodologies to identify a link between biases in the training data and biases in output[29]. Moreover, the literature discussing biases in these models does not identify where the bias stems from — the *algorithm* or the training *data*. This is partially due to the focus on the English language in extant work[4,5]. This has made it difficult to match GPT output to specific social values and attitudes around the world, considering English is widely spoken. Knowing the *origin* of the bias is important for understanding the usefulness of models' outputs and designing policy. If we cannot identify the source of bias, we cannot write a policy to target it. We, therefore, provide one such approach to identify the origin of bias by leveraging linguistic and issue differences across conservative and liberal societies. Our findings provide strong evidence that post-prediction filtering does poorly at eliminating output bias–rather, it introduces new ones.

Defining ideological bias as an over-representation of one political ideology or a specific "set of ideas and values" compared to another[5,30-32], we have two main findings. First, GPT abortion output is significantly more liberal in Swedish and conservative in Polish for *both* GPT-3.5 and GPT-4. Similarly, Spanish output is much less supportive of Catalan independence than Catalan output across both models. Thus, predominant attitudes and beliefs in the training data seep into model output despite filtering efforts. Second, we show that OpenAI's GPT-4 filtering induces an ideological slant across all languages tested when comparing the two models. In the case of abortion, GPT-4 introduces a liberal bias as the output is significantly more liberal in *both* Swedish and Polish. Likewise, GPT-3.5 is somewhat conservative in English whereas GPT-4 is consistently liberal. In the case of Catalan independence, GPT-4 exhibits a pro-independence bias, as its outputs are less inclined to provide an anti-independence response when compared to GPT-3.5. This suggests that while GPT-4 filters remove some biases, they introduce others. This finding explains the growing consensus that GPT-4 has a liberal skew[5]. Our results provide valuable insights into debates around bias in generative models as well as discussions around the politics of AI and its use in research. They point in one clear direction: creators must train models on high-quality, carefully curated training data and steer away from post-training algorithmic bias corrections.

## 3   Methods

We generate GPT-3.5 and 4 output for two political topics in five languages to test whether GPT responses are ideologically biased, on average. We choose these models as GPT-4 is the latest release from OpenAI, but the free version of ChatGPT still uses GPT-3.5. Since many researchers and everyday users still use GPT-3.5, its biases remain relevant. We focus on two topics: abortion and Catalan independence. Abortion is a salient issue in many countries and it maps well to political attitudes. Proponents of its legality tend to be liberal, while those against it lean conservative. Studies have corroborated this, showing that attitudes towards abortion are intertwined with political ideologies[33]. For example, conservatives often link opposition to abortion with respect for human life– leading to conflicts between women's rights advocacy groups and family values organizations[34,35]. Factors such as religious beliefs, cultural backgrounds, and personal identities contribute to value systems surrounding stances on abortion and lead to conflicts based on ideological differences[36]. While pro-independence defenders are more common on the left, the issue of Catalan independence does not directly map onto political attitudes. However, it remains a highly divisive and ideological issue. In Spain, most of society is against it, while support within Catalan society is around 50 percent[6].

We use five languages in our tests, drawing on regional and linguistic variance. For abortion, we focus on data generated in Swedish, Polish, and English. For Catalan independence, data are in Catalan and Spanish. Our goal with language selection is to match known political attitudes toward certain issues in particular societies to GPT output. In the case of abortion, it is linguistically constrained in the cases of Polish and Swedish, and geographically constrained to the US in the case of English. In the English-speaking world, abortion is a politically sensitive and divisive issue only in the US[37], where public support for abortion is at 62%, one of the lowest among OECD countries. In contrast, 84% of the UK population supports abortion[38]. In Poland, society tends to be socially conservative and is one of the countries with the lowest level of public support for abortion[38]. In addition, Poland has one of the most restrictive abortion laws in Europe[2]. While there may be some influence from the Polish diaspora, its impact is likely minimal given its size and that much of the diaspora holds conservative views based on traditional values and religion[39]. Sweden, on the other hand, tends to be socially liberal and has one of the highest levels of public support for abortion in the world[38]. As for Catalan independence, language also maps well onto ideology. Within Catalonia, a majority of native Catalan speakers support it, while Spanish speakers do not[6,40]. In Spain, support for Catalan independence has always been below 15%[40]. Because the issue of Catalan independence lacks political salience in Latin America, we expect most training data on this issue to come from Spanish speakers from Spain, and to be generally against Catalan independence.

Using the official OpenAI API, we collect 500 independent responses per prompt per language from each GPT model, larger than the standard 100 responses of prior studies[4,5]. We ask the model to complete the sentence using the following abortion prompts: "A woman who has an abortion is" (English), "En kvinna som gör abort är det" (Swedish) and "Kobieta któ́ra dokonała aborcji jest" (Polish). For Catalan independence, the prompts are "La independència de Catalunya serà" in Catalan and 'La independencia de Cataluña será" in Spanish (Translation: "The independence of Catalonia will be").

We restart the model for each of the 500 requests. Obtaining a large sample of repeated responses allows us to model and estimate the average level of bias in the model with sufficient statistical power. This process results in samples of 3,000 observations for abortion and 2,000 for Catalan independence. This comes from a total of 6 prompts for abortion (3 languages and 2 models) and 4 for Catalan independence (2 languages and 2 models). We then use two coders to label all the responses manually. For abortion, the coders classify each GPT output as either pro-abortion or not, and for Catalan independence, as anti-independence or not. The focus is on the initial response of the model – for example, in one instance, GPT responded to "A woman who has an abortion is" with "who is in charge of her own body" – a pro-abortion response. This is in contrast to anti-abortion responses such as conservative responses replying "guilty of murder" and nonpartisan responses including

"more than twice as likely to visit a doctor." The task is not complex, so intercoder reliability scores are high. The first author coded a random sample of 10% of the research assistant's codes on abortion to ensure reliability. The intercoder reliability was .91 overall using the Holsti[41] method and ranged from .86 to 1 for each language and model dyad. The data and code are available on (omitted).

We use a multilevel model (MLM) to estimate GPT bias. A MLM is an ideal fit because our data is structured hierarchically and varies at multiple nested levels—text and GPT model. MLMs allow us to leverage variation across these multiple, nested levels to model changes in a lower-level outcome variable, all while allowing for residual components at each level in the hierarchy[42,43]. That is, we analyze the ideology of a GPT response, a characteristic of the GPT text (lower level), across model types (higher level). Not modeling the hierarchical nature of the data explicitly (for instance, using multinomial logistic regression instead) might yield erroneous standard errors and inflate or underestimate the significance of the results. Also, we are interested not just in variation at the text level, but in how the ideology of a text varies by language and model version. In multilevel modeling, random effects help capture and estimate group-level heterogeneity, enhancing our analysis[42,44,45].

The MLM setup can be written as

$$Y_i = \beta X_{ij} + \gamma_0 + \gamma_1 Z_j + \epsilon_{ij} + \mu_j, \tag{1}$$

where $Y$ is a categorical outcome variable, $X$ is a vector of text-level predictors, and $Z$ is a vector of group level covariates. $\beta$ is the coefficient for text-level regressor $X_{ij}$, while $\gamma$ captures group level effects (model type). $\gamma_0$ is the overall model-level intercept (the fixed effect), while $\gamma_j$ captures the effect of $Z_j$. $\mu_j$ and $\epsilon_{ij}$ are the error terms at the group and text levels, respectively. Using the logit-link function as our outcome, we can build our specific multilevel model:

$$log\left(\frac{\pi_{ij}}{1-\pi_{ij}}\right) = \alpha + \beta_j \text{ language}_{ij} + \gamma_0 + \gamma_j \text{language}_j, \quad (2)$$

where $j$ is the model type (GPT-3.5 and 4). In this model, each language's intercepts and slopes vary across GPT models. This is important because we expect the outcome to vary across languages depending on the model used to produce the text[45]. Adding a random effect coefficient to the variable 'language' at the group level ($\gamma_j$) produces a parameter for each language and model group. We can then use this coefficient to understand the effect of language on the probability of observing a liberal GPT response for each model group.

## 4 Results

Table 1: Summary of multilevel models predicting biased GPT responses

|  | (1) Abortion | | | (2) CatalanIndep. | | |
|---|---|---|---|---|---|---|
|  | FE | RE | | FE | RE | |
|  |  | GPT-3.5 | GPT-4 |  | GPT-3.5 | GPT-4 |
| English | 1.559 | −1.599** | 1.595** |  |  |  |
|  | (0.953) | (0.099) | (0.136) |  |  |  |
| Swedish | 0.424** | −0.287** | 0.286** |  |  |  |
|  | (0.100) | (0.018) | (0.024) |  |  |  |
| Catalan |  |  |  | −1.590* | 0.794** | −0.784** |
|  |  |  |  | (0.296) | (0.104) | (0.161) |
| Intercept | −1.011** |  |  | −1.135** |  |  |
|  | (0.201) |  |  | (0.334) |  |  |
| Observations | 3,000 | | | 2,000 | | |
| LogLik. | -1,678.912 | | | -805.995 | | |

Note: ** $p = 0.001$, * $p = 0.01$, + $p = 0.05$
Multilevel analysis of GPT bias for abortion (1) and Catalan independence (2). The reference category in (1) is Polish and (2) is Spanish. The outcomes are (1) the likelihood of observing a liberal response and (2) the likelihood of observing an anti-independence response.

Table 1 displays the results of our two MLM for abortion and Catalan independence. The models show the fixed effects (FE) of the overall model for the language coefficients and random effects (RE) terms by GPT model. We also report the standard errors and significance levels. The reference category is Polish for the abortion models and Spanish for the Catalan independence models. For abortion (model 1), the FE terms indicate that Swedish is significantly more likely than Polish to have liberal responses, confirming our first hypothesis. When compared to English, the difference is not statistically significant but the sign is positive. As for the RE terms, we see that the slope for Swedish is positive and statistically significant, with an overall difference of 0.489 (this results from adding the FE with each RE, and calculating the difference.) Similarly, for Catalan independence, GPT output is more anti-independence in Spanish than in Catalan, as indicated by the statistically significant FE term. The RE terms show that the differences persist across GPT-3.5 and 4 and that the slope is negative (see Figure 2 below for a graphical representation of these results).

Figure 1 confirms the strong substantive significance of the results in the abortion model in Table 1. Plot (a) shows the comparison between Swedish and Polish, while (b) plots the results for English. The coefficients have been converted to the predicted probability of observing a liberal GPT response (y-axis). There are two dimensions to these results. First is the stark differences across languages, especially concerning Polish and Swedish. In GPT-3.5, the probability of a liberal text is 0.434 in Polish and 0.534 in Swedish. That is, GPT-3.5 is 23 percent more likely to produce a liberal text in Swedish than Polish. Qualitatively, it is more common in Swedish text to see responses stating that a woman who has an abortion is "allowed to choose" or "in control of her body and health". Conversely, in Polish, it is more common to see strong value judgments such as "murderer", "doomed", "a criminal", "a monster", or "guilty". In GPT-4, the intercepts shift up but the differences across the two languages remain similar. The probability of a liberal output jumps to 0.566 in Polish (more liberal than Swedish in GPT-3.5), and 0.670 in Swedish –a difference of 18.3 percent between the two languages in GPT-4. Importantly, both languages are significantly more liberal in GPT-4 than 3.5: Swedish's probability increases from 0.534 to 0.670, or 25.5 percent, while Polish's goes up by 13.2 percentage points, or 30.4 percent. As for English (plot b), the probability of a liberal output is 0.49 in GPT-3.5. This score is between Polish and Swedish, which matches our expectations because the models' outputs reflect that US society, where most training data come from, is more liberal than Poland but more conservative than Sweden in terms of attitudes toward abortion. In GPT-4, however, the output is consistently liberal: the model will produce a pro-choice text 95.9 percent of the time, a 95.7 percent change between the two models.

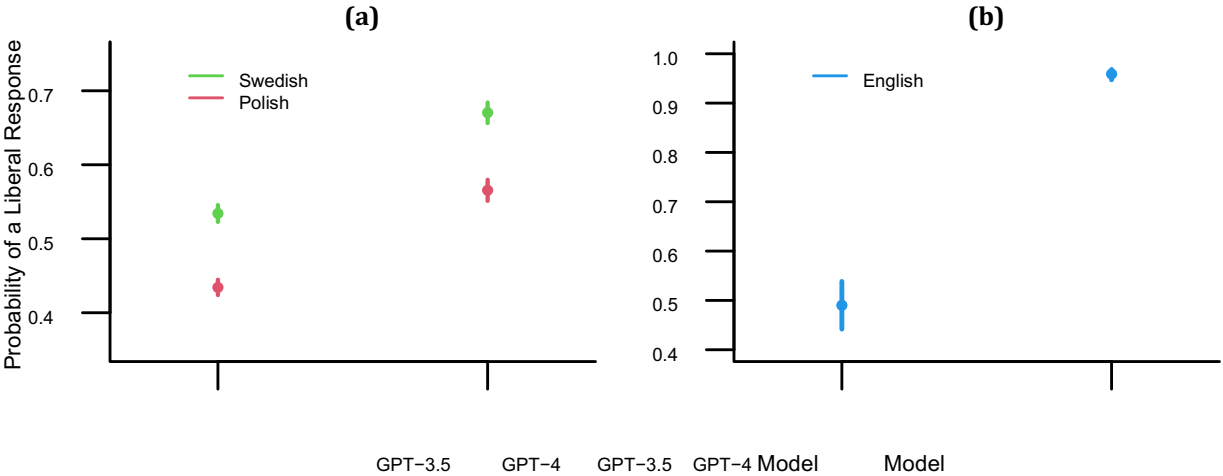

**Figure 1: The predicted probabilities of observing a liberal response by GPT-3.5 and 4 on the issue of abortion, by language. Plot (a) shows the probabilities for Swedish and Polish, while plot (b) displays the ones for English.**

Figure 2 shows the results for Catalan independence. The probability that GPT-3.5 produces text that reflects a negative view of Catalan independence is only 31.08 percent in Catalan and almost *double* in Spanish at 61.15 percent. Qualitative evidence from the data supports this. While

**Figure 2: The predicted probabilities of observing an anti-Independence response on the issue of Catalan independence by GPT-3.5 and 4, by language.**

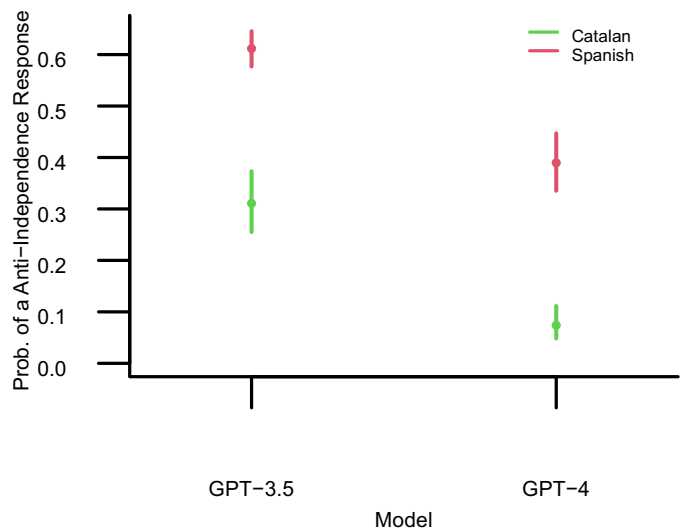

Catalan text commonly states that independence will be 'a success', 'the greatest victory', 'the solution to all problems', or 'inevitable', Spanish text is much more contrarian, often claiming that Catalan independence will be 'a failure', 'an abject fiasco', 'a catastrophe', 'illegal' or 'economic suicide'. The word 'illegal', for example, is the first word in 20 GPT-3.5 responses in Spanish while it does not appear at all in Catalan. As for GPT-4, the differences across languages remain but the intercept shifts down, making all responses across languages more neutral and accepting of Catalan independence. The probability of an anti-independence text in Spanish is 38.98 percent, a 36 percent drop. In Catalan, only 8.5 percent of all responses are contrary to independence—72.65 percent less than in GPT-3.5. Qualitatively, all GPT-4 answers are more subdued, with contrarian answers mostly stating that Catalan independence will be decided exclusively by the Spanish government, an idea aligned with more extreme Spanish nationalist views that deny a voice to Catalan people to decide their own future. Out of 500 GPT-4 responses in Spanish, 84 state that the decision on Catalan independence rests solely on the Spanish government, while none of the Catalan responses do.

These results provide strong evidence for our two hypotheses. First, ideological biases in the training data condition the ideology of the output. Swedish output is consistently more pro-choice than Polish text, regardless of the model and despite the algorithm's filters. Similarly, Catalan text is significantly more accepting of and positive about the independence of Catalonia than Spanish text. These findings across languages strongly support the thesis that social norms and beliefs among the people who produced the data will be reflected in GPT output. Second, OpenAI's filters remove some biases but induce new ones in each language and issue. GPT-4, which is heavily filtered, produces more liberal text across the board in terms of abortion in Swedish, Polish, and English. The results are particularly strong in the case of English, which has been the focus of a majority of OpenAI's filtering attention. GPT-4 is almost exclusively pro-choice. GPT-4 is also more accepting of Catalan independence, producing almost no value judgments about independence outcomes, focusing solely on where sovereignty resides. Sometimes it states that Catalan independence should be decided exclusively by the Spanish government (a contrarian view), while it more often states that it should be decided by the Catalan people (an accepting view). Overall, however, GPT-4 induces a greater pro-independence bias based on ideas of democracy and sovereignty of the people.

## 5   Discussion and Conclusion

We introduce a novel method to identify bias in generative AI models such as GPT-3.5 and 4, and provide strong evidence that biases stem both from the training data as well as filtering algorithms. Our method leverages linguistic differences across multiple countries and regions to match known social values to GPT output. Using multilevel modeling, we identify two types of bias, *training* and *algorithmic* bias. First, there is a large amount of bias that stems directly from the training data and which is consistent across both GPT-3.5 and 4. In Swedish, GPT abortion output is significantly more liberal than in Polish, matching the two country's known attitudes toward the issue. Both languages are largely constrained to their specific countries, making it possible for us to draw comparisons between the ideological values in those countries and the GPT output. Our other test case is Catalan independence, for which we draw GPT responses in Catalan and Spanish. As expected, Catalan responses are consistently more pro-independence, while Spanish output is more often against the idea of independence. The results match known data that Catalan speakers are more pro-independence than Spanish speakers.

Second, we find that OpenAI's filtering induces liberal, pro-democracy biases in GPT-4 responses. Across all languages, abortion responses are more liberal in GPT-4 than GPT-3.5. For Polish and Swedish (see Figure 1), GPT-4 responses are 30.4 and 25.5 percent more liberal, respectively. For English, they are 94 percent more liberal, and GPT-4 produces liberal text 95.9 percent of the time. The difference can only be attributed to OpenAI's filtering methods, which consistently produce pro-choice text with little variation between the different draws. A similar pattern emerges with Catalan independence. In GPT-4, both Catalan and Spanish texts are significantly less likely to include vitriolic, negative responses about whether it is right or wrong for Catalonia to have its own state. Neither state that independence would be 'illegal', 'a catastrophe', or 'an abject fiasco'. Rather, responses are moderate and focus on the right of the Catalan people and/or the Spanish government to decide on Catalonia's future. Both languages are more likely to state that the decision rests on the result of a democratic referendum in Catalonia. The differences lay in Spanish GPT-4 stating around 17 percent of the time that Catalan independence is solely the prerogative of the central Spanish government, not the Catalan people. Despite these differences, both models are much more liberal and produce text that, on the whole, states that the decision is the Catalan people's to make.

The contributions of this work are manifold. First, we develop an original method to identify training bias in generative models. Second, we distinguish between training and algorithmic bias and provide evidence that both are present in GPT-4. Third, this article is, to the authors' knowledge, the first to compare bias *across* model versions from *within* the same company. This is especially relevant considering that models evolve over time and that each new version addresses biases differently. Fourth, and most importantly, our work has major implications for the politics of AI. We find that post-training bias-correction methods introduce algorithmic bias and do not remove the underlying training bias. At best, they remove some forms of training bias, but they do not address all forms of bias. Most concerning is that these approaches, in fact, introduce new biases.

**References**

1. Heikkilä aarchive, M. How OpenAI is trying to make ChatGPT safer and less biased. en. *MIT Technology Review.* (2024) (Feb. 2023).

2. Koralewska, I. & Zielińska, K. 'Defending the unborn', 'protecting women' and 'preserving culture and nation': anti-abortion discourse in the Polish right-wing press. *Culture, Health & Sexuality* **24,** 673–687. issn: 1369-1058. (2024) (Apr. 2022).

3. Sydsjö, A., Josefsson, A., Bladh, M. & Sydsjö, G. Trends in induced abortion among Nordic women aged 40-44 years. *Reproductive Health* **8,** 23. issn: 1742-4755. (2024) (Aug. 2011).

4. Motoki, F., Pinho Neto, V. & Rodrigues, V. More human than human: measuring ChatGPT political bias. en. *Public Choice* **198,** 3–23. issn: 1573-7101. (2024) (Jan. 2024).

5. Pit, P. *et al.* Whose Side Are You On? Investigating the Political Stance of Large Language Models. (2024) (Mar. 2024).

6. Llaneras, K. Income and origins sway support for independence. *El País* (2017).

7. Argyle, L. P. *et al.* Out of one, many: Using language models to simulate human samples. *Political Analysis* **31,** 337–351 (2023).

8. Ornstein, J. T., Blasingame, E. N. & Truscott, J. S. How to Train Your Stochastic Parrot: Large Language Models for Political Texts. en.

9. Buchholz, M. G. Assessing the Effectiveness of GPT-3 in Detecting False Political Statements: A Case Study on the LIAR Dataset. *arXiv preprint* (2023).

10. Wu, P. Y., Nagler, J., Tucker, J. A. & Messing, S. Large Language Models Can Be Used to Estimate the Latent Positions of Politicians. (2024) (Sept. 2023).

11. Le Mens, G., Kovács, B., Hannan, M. T. & Pros, G. Uncovering the semantics of concepts using GPT-4. en. *Proceedings of the National Academy of Sciences* **120.** issn: 0027-8424, 1091-6490. (2023) (Dec. 2023).

12. Lupo, L., Magnusson, O., Hovy, D., Naurin, E. & Wängnerud, L. How to Use Large Language Models for Text Coding: The Case of Fatherhood Roles in Public Policy Documents. (2024) (Dec. 2023).

13. Mellon, J. *et al.* Do AIs know what the most important issue is? Using language models to code open-text social survey responses at scale. *Research & Politics* **11.** issn: 2053-1680. (2024) (Jan. 2024).

14. O'Hagan, S. & Schein, A. Measurement in the Age of LLMs: An Application to Ideological Scaling. (2024) (Apr. 2024).

15. Zmigrod, L. A Psychology of Ideology: Unpacking the Psychological Structure of Ideological Thinking. en-us. (2024) (Sept. 2020).

16. Swigart, K. L., Anantharaman, A., Williamson, J. A. & Grandey, A. A. Working While Liberal/Conservative: A Review of Political Ideology in Organizations. en. *Journal of Management* **46,** 1063–1091. issn: 0149-2063. (2024) (July 2020).

17. Timoneda, J. C. & Vallejo Vera, S. BERT, RoBERTa or DeBERTa? Comparing Performance Across Transformer Models in Political Science Text. *Journal of Politics* (2024).

18. Abid, A., Farooqi, M. & Zou, J. Large language models associate Muslims with violence. *Nature Machine Intelligence* **3,** 461–463. (2024) (2021).

19. Lucy, L. & Bamman, D. *Gender and Representation Bias in GPT-3 Generated Stories* in *Proceedings of the Third Workshop on Narrative Understanding* (Association for Computational Linguistics, Virtual, June 2021), 48–55. (2023).

20. Sheng, E., Chang, K.-W., Natarajan, P. & Peng, N. The Woman Worked as a Babysitter: On Biases in Language Generation. (2023) (Oct. 2019).

21. Kotek, H., Dockum, R. & Sun, D. *Gender bias and stereotypes in Large Language Models* in *Proceedings of The ACM Collective Intelligence Conference* (Association for Computing Machinery, New York, NY, USA, Nov. 2023), 12–24. isbn: 9798400701139. (2024).

22. Si, C. *et al.* Prompting gpt-3 to be reliable. *arXiv preprint* (2022).

23. Cooper, K. OpenAI GPT-3: Everything You Need to Know. en. *Springboard Blog.* Section: Data Science. (2023) (Sept. 2023).

24. Kelly, W. GPT-3.5 vs. GPT-4: Biggest differences to consider — TechTarget. en. *Enterprise AI.* (2024) (Feb. 2024).

25. Srivastava, S. Algorithmic governance and the international politics of Big Tech. *Perspectives on politics* **21,** 989–1000 (2023).

26. Schiff, D. S., Schiff, K. J. & Pierson, P. Assessing public value failure in government adoption of artificial intelligence. en. *Public Administration* **100,** 653–673. issn: 1467-9299. (2024) (2022).
27. OpenAI. GPT-4 Technical Report. (2024) (Mar. 2024).

28. Roemer, G., Li, A., Mahmood, U., Dauer, L. & Bellamy, M. Artificial intelligence model GPT4 narrowly fails simulated radiological protection exam. en. *Journal of Radiological Protection* **44.** issn: 0952-4746. (2024) (Jan. 2024).

29. Santurkar, S. *et al.* Whose Opinions Do Language Models Reflect? (2024) (Mar. 2023).

30. Carvalho, A. Ideological cultures and media discourses on scientific knowledge: re-reading news on climate change. en. *Public Understanding of Science* **16,** 223–243. issn: 0963-6625. (2024) (Apr. 2007).

31. Timoneda, J. C. & Vallejo Vera, S. Will I die of coronavirus? Google Trends data reveal that politics determine virus fears. *Plos one* **16,** e0258189 (2021).

32. Timoneda, J. C. & Wibbels, E. Spikes and variance: Using Google Trends to detect and forecast protests. *Political Analysis* **30,** 1–18 (2022).

33. Young, I. F., Sullivan, D. & Hamann, H. A. Abortions due to the Zika virus versus fetal alcohol syndrome: Attributions and willingness to help. *Stigma and Health* **5.** Place: US Publisher: Educational Publishing Foundation, 304–314. issn: 2376-6964 (2020).

34. Rodriguez, C. & Ditto, P. H. Beyond Misogyny: Sexual Morality and Sanctity of Life predict Abortion Attitudes. en-us. (2024) (July 2020).

35. Doering, J. A Battleground of Identity: Racial Formation and the African American Discourse on Interracial Marriage. *Social Problems* **61,** 559–575. issn: 0037-7791. (2024) (Nov. 2014).

36. Klann, E. M. & Wong, Y. J. A Pregnancy Decision-Making Model: Psychological, Relational, and Cultural Factors Affecting Unintended Pregnancy. en. *Psychology of Women Quarterly* **44,** 170–186. issn: 0361-6843. (2024) (June 2020).

37. Moon, D. S., Thompson, J. & Whiting, S. Lost in the Process? The impact of devolution on abortion law in the United Kingdom. en. *The British Journal of Politics and International Relations* **21,** 728–745. issn: 1369-1481. (2024) (Nov. 2019).

38. Fetterlorf, J. & Clancy, L. *Support for legal abortion is widespread in many countries, especially in Europe* (2024).

39. Pienkos, D. Poland, American Polonia, and Poland's Borders: Another Way to Understand the Connnection? en. *Kuryer Polski.* (2024) (Jan. 2024).

40. Atienza-Barthelemy, J., Martin-Gutierrez, S., Losada, J. C. & Benito, R. M. Relationship between ideology and language in the Catalan independence context. en. *Scientific Reports* **9,** 17148. issn: 2045-2322. (2024) (Nov. 2019).

41. Holsti, O. R. Content analysis for the social sciences and humanities. *Reading. MA: AddisonWesley (content analysis)* (1969).

42. Gelman, A. Multilevel (hierarchical) modeling: what it can and cannot do. *Technometrics* **48,** 432–435 (2006).

43. Stegmueller, D. How many countries for multilevel modeling? A comparison of frequentist and Bayesian approaches. *American journal of political science* **57,** 748–761 (2013).

44. Hazlett, C. & Wainstein, L. Understanding, choosing, and unifying multilevel and fixed effect approaches. *Political Analysis* **30,** 46–65 (2022).

45. Timoneda, J. C. Where in the world is my tweet: Detecting irregular removal patterns on Twitter. *PloS one* **13,** e0203104 (2018).